\documentclass[conference]{IEEEtran}
\IEEEoverridecommandlockouts
\usepackage{cite}
\usepackage{amsmath,amssymb,amsfonts}
\usepackage{algorithmic}
\usepackage{graphicx}
\usepackage{textcomp}
\usepackage{xcolor}
\def\BibTeX{{\rm B\kern-.05em{\sc i\kern-.025em b}\kern-.08em
    T\kern-.1667em\lower.7ex\hbox{E}\kern-.125emX}}
\usepackage[nolist]{acronym}
\usepackage{svg}
\usepackage{hyperref}
\usepackage{booktabs}

\begin{document}
\newacro{mae}[MAE]{mean absolute error}
\newacro{der}[DER]{distributed energy resources}
\newacro{emt}[EMT]{electromagnetic transient}
\newacro{ml}[ML]{machine learning}
\newacro{fc}[FC]{fault classification}
\newacro{fl}[FL]{fault localization}
\newacro{pr}[PR]{protection relay}
\newacro{res}[RES]{renewable energy sources}

\newacro{bc}[BC]{Bagging Classifier}
\newacro{dt}[DT]{Decision Tree}
\newacro{et}[ET]{Extra Trees}
\newacro{gb}[GB]{Histogram-based Gradient Boosting}
\newacro{knn}[KNN]{K-Nearest Neighbors}
\newacro{lg}[LG]{Logistic Regression}
\newacro{mlp}[MLP]{Multi-Layer Perceptron}
\newacro{rf}[RF]{Random Forest}
\newacro{ridge}[Ridge]{Ridge Regression}
\newacro{sgd}[SGD]{Stochastic Gradient Descent}
\newacro{stacking}[Stacking]{Stacking Ensemble}
\newacro{svc}[SVC]{Support Vector Classifier}
\newacro{svm}[SVM]{Support Vector Machine}
\newacro{svr}[SVR]{Support Vector Regressor}
\newacro{voting}[Voting]{Voting Ensemble}
\newacro{lstm}[LSTM]{Long Short-Term Memory}

\title{Controlled Comparison of Machine Learning Models for Fault Classification and Localization in Power System Protection}
\author{
\IEEEauthorblockN{
Julian Oelhaf\textsuperscript{1}\textsuperscript{*},
Georg Kordowich\textsuperscript{2},
Changhun Kim\textsuperscript{1},
Paula Andrea Pérez-Toro\textsuperscript{1},
Christian Bergler\textsuperscript{3}\\
Andreas Maier\textsuperscript{1},
Johann J\"ager\textsuperscript{2},
Siming Bayer\textsuperscript{1}
}

\IEEEauthorblockA{
\textit{\textsuperscript{1}Pattern Recognition Lab, Friedrich-Alexander-Universit\"at Erlangen-N\"urnberg} \\
\textit{\textsuperscript{2}Institute of Electrical Energy Systems, Friedrich-Alexander-Universit\"at Erlangen-N\"urnberg} \\
\textit{\textsuperscript{3}Department of Electrical Engineering, Media and Computer Science, Ostbayerische Technische Hochschule Amberg-Weiden} \\
{\textsuperscript{*}Corresponding author: julian.oelhaf@fau.de}
}
}

\maketitle

\begin{abstract}
The increasing complexity of modern power systems, driven by the integration of inverter-based and distributed energy resources, challenges the reliability of conventional protection schemes and motivates the use of machine learning for protection tasks. However, published results are often difficult to compare because datasets, sensing assumptions, and decision horizons vary across studies. This paper presents a controlled comparison of machine learning models for fault classification (FC) and fault localization (FL) under identical sensing, timing, and validation conditions on a common electromagnetic transient dataset, using decision windows of 10--50\,ms to reflect protection-relevant time scales. For FC, the best-performing nonlinear models achieve F1 scores above 0.98 already at 10\,ms, while lower-capacity models degrade at shorter horizons but improve with longer windows, indicating that relevant fault-type information is already present in the earliest transient. For FL, the top-performing models reach a stable localization error of about 10\% of normalized line length across all evaluated horizons, while weaker models form a clearly separated second performance tier. Line-resolved analysis shows that localization accuracy varies across grid segments, indicating topology-dependent difficulty rather than insufficient temporal context alone. These findings provide a controlled reference for comparing machine learning models across two protection tasks with fundamentally different information requirements.
\end{abstract}

\begin{IEEEkeywords}
Power system protection, fault classification, fault localization, 
machine learning, electromagnetic transient simulation, 
task-dependent evaluation
\end{IEEEkeywords}

\section{Introduction}
\label{sec:intro}

The transition towards decentralized power systems, driven by the integration of \ac{res} and \ac{der}, reshapes grid operation. Increasing shares of inverter-based generation and the adoption of hybrid AC--DC architectures~\cite{protection_and_automation_b5_protection_2015} expand the spectrum of operating and fault scenarios in modern grids~\cite{vde_zellulare_2015}. These include highly meshed topologies, multi-terminal arrangements, and dynamic operational strategies such as curative redispatch with temporary overloads exceeding nominal conditions.

These developments challenge conventional protection systems, which rely on deterministic algorithms with fixed thresholds and static models~\cite{blackburn_protective_2014}. In standard operation, protection must remain inactive, while in fault conditions -- such as short circuits, ground faults, conductor breaks, or thermal overloads -- it must act immediately and selectively through circuit breakers. However, the variability and uncertainty introduced by \ac{res} and \ac{der} increasingly blur the distinction between normal and faulty states~\cite{vde_zellulare_2015}.

Short-circuits impose significant thermal and mechanical stress on grid assets, while \ac{der} inject fault currents with magnitudes and waveforms unlike synchronous machines, altering fault signatures and complicating selective protection~\cite{protection_and_automation_b5_protection_2015}. These developments motivate data-driven approaches: \ac{ml} methods can capture nonlinear dynamics in voltage and current waveforms and have shown promising results for fault detection and line identification~\cite{abdullah_ultrafast_2018}, though their reliability remains limited by data quality, class imbalance, and limited interpretability in safety-critical settings~\cite{oelhaf_impact_2025}, motivating consistent evaluation to distinguish genuine task limitations from dataset or model artefacts.

A recent scoping review~\cite{oelhaf_scoping_2025} systematically analyzed \ac{ml} applications in power system protection across diverse grid types and revealed substantial inconsistencies in simulation setups, preprocessing strategies, and evaluation metrics, making meaningful comparison difficult. Most works address \ac{fc} or \ac{fl} in isolation and under heterogeneous conditions. Jones~et~al.~\cite{jones_machine_2021} evaluate \ac{fc} and coarse fault-region identification with a single \ac{svm} on a distribution feeder, hinting at a task-level asymmetry but without controlling for decision horizon or model diversity. Harish~et~al.~\cite{harish_comparative_2023} compare multiple architectures for \ac{fc} on a transmission system yet do not address \ac{fl} at all. Taheri~et~al.~\cite{taheri_fault-location_2021} show that embedding physical priors into an \ac{lstm} reduces \ac{fl} error on parallel lines, but study only a single model and a single task. Without a shared evaluation reference, reported results are difficult to interpret: an F1 score of 99\,\% on a simple radial feeder and 90\,\% on a meshed topology reflect entirely different task difficulties, yet the literature routinely treats such figures as comparable.

To the best of our knowledge, this is the first controlled comparison of machine learning models for \ac{fc} and \ac{fl} under strictly identical sensing, timing, and validation conditions across a diverse set of model families. Building on prior work that evaluated fault detection and line identification under unified conditions~\cite{oelhaf_systematic_2025}, this study extends the analysis to two tasks that share the same measurements but differ fundamentally in their information requirements. By fixing all evaluation assumptions independently of the learning model, we isolate task-dependent performance limits from model-specific effects and reveal a clear asymmetry: \ac{fc} performance is largely insensitive to decision horizon for high-capacity models, with the discriminative fault signature present already in the shortest evaluated window, while \ac{fl} exhibits a persistent error floor stable across all top-performing models and window lengths, consistent with structural identifiability limits imposed by the grid topology rather than insufficient model 
capacity.

\begin{figure*}[t]
    \centering
    \includegraphics[width=0.85\linewidth]{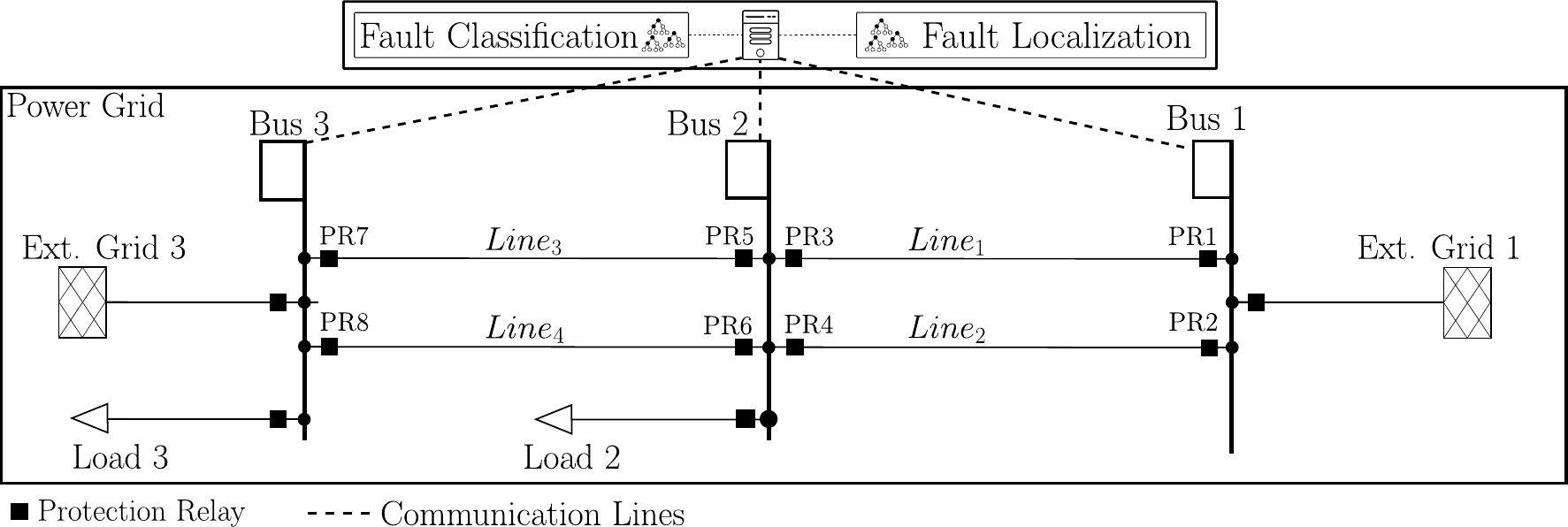}
    \caption{Double Line grid topology used for data generation and experiments. \ac{emt} simulations compute instantaneous \(V\) and \(I\) to capture transients; dataset settings: 90\,kV nominal voltage, 6400\,Hz sampling, 1\,s episodes. (Adapted from~\cite{kordowich_protect-90_2026})}
    \label{fig:double_line_grid_topology}
\end{figure*}

\section{Methodology}
\label{sec:methodology}

This section describes the dataset and grid topology, signal processing pipeline, task formulations, and model evaluation protocol used in this study.

\begin{table}[t]
\centering
\caption{Parameter ranges used for domain randomization in the PROTECT-90 dataset~\cite{kordowich_protect-90_2026}.}
\label{tab:param_var_overview}
\begin{tabular}{l l c c}
\toprule
\textbf{Component} & \textbf{Parameter} & \textbf{Min} & \textbf{Max} \\
\midrule
Line & Length (km) & 10 & 60 \\
     & Reactance $X'$ ($\Omega$/km) & 0.35 & 0.45 \\
     & Resistance $R'$ ($\Omega$/km) & 0.01 & 0.20 \\
     & Capacitance $C'$ (nF/km) & 8.5 & 10 \\
\midrule
Load & Active power $P$ (MW) & 20 & 50 \\
     & Reactive power $Q$ (MVar) & $-20$ & 20 \\
\midrule
Grid & Short-circuit power $S_k''$ (MVA) & 90 & 1000 \\
     & Voltage setpoint $V_{\text{set}}$ (pu) & 0.95 & 1.05 \\
     & Phase angle $\phi$ (deg) & $-180$ & 180 \\
\bottomrule
\end{tabular}
\end{table}

\subsection{Dataset and Grid Topology}\label{sec:dataset}

The evaluation uses the publicly available PROTECT-90 dataset~\cite{kordowich_protect-90_2026}, based on a standard ``Double Line'' topology commonly used in protection studies~\cite{ziegler_numerical_2011}. Although this topology does not capture the effects of large-scale meshed networks, it is widely used as a benchmark for assessing protection difficulty. In contrast to simple radial feeders, where fault distance can often be inferred directly, the ``Double Line'' topology exhibits structural symmetries, including intermediate infeeds and currents on parallel lines, that substantially complicate fault identification. It therefore provides a conservative lower bound on achievable performance and is particularly well-suited for isolating task-dependent differences between \ac{fc} and \ac{fl}. The grid topology is shown in Fig.~\ref{fig:double_line_grid_topology}.

All waveforms are generated in DIgSILENT PowerFactory using the \ac{emt} simulation module. \Ac{emt} simulation is chosen over RMS simulation because it captures fast transient dynamics, including high-frequency components and waveform distortion at sub-cycle timescales, that are critical for protection-relevant fault signatures within the evaluated decision horizons of 10--50\,ms.

The dataset comprises 9022 simulation episodes of 1\,s duration at 6400\,Hz, each representing a unique grid and fault configuration obtained through domain randomization of line parameters, loading conditions, fault locations, and external grid settings (Table~\ref{tab:param_var_overview}). Parameter ranges are drawn from established reference values~\cite{oeding_elektrische_2016,roeper_kurzschlusstrome_1984} and span physically plausible high-voltage operating conditions without introducing numerically unstable configurations. PROTECT-90 covers single-phase-to-ground, line-to-line, double-line-to-ground, and three-phase short circuits at 90\,kV, balanced across fault types and approximately uniformly distributed across fault locations and phases; full details are provided in~\cite{kordowich_protect-90_2026}.

Three-phase voltage and current are recorded jointly at each \ac{pr}, as single-phase measurements are insufficient to distinguish fault types requiring cross-phase comparison -- in particular phase-to-phase and double-line-to-ground faults:

\begin{equation*}
\begin{aligned}
    I_{PR}(t) &= [I_A(t), I_B(t), I_C(t)], \\
    V_{PR}(t) &= [V_A(t), V_B(t), V_C(t)], \quad t \in [0, 1]\,\text{s}
\end{aligned}
\end{equation*}

These signals form a multivariate time series per relay:

\begin{equation*}
    X_{PR}(t) = \begin{bmatrix} I_{PR}(t) \\ V_{PR}(t) \end{bmatrix}
\end{equation*}

with subscripts \( A \), \( B \), and \( C \) denoting the phases.

\begin{table}[t]
\centering
\caption{
Decision window lengths, corresponding number of discrete samples (time steps) per window at $f_s = 6400\,\mathrm{Hz}$, total number of extracted windows, and number of fault-containing windows.
}

\label{tab:window_length_overview}
\begin{tabular}{c c c c}
\toprule
\textbf{Window} & \textbf{Samples} & \textbf{\# Windows} & \textbf{\# Fault} \\
\textbf{length} & \textbf{/ window} &  & \textbf{windows} \\
\midrule
10\,ms  & 64  & 279\,682 & 9\,022  \\
20\,ms  & 128 & 261\,638 & 27\,066 \\
30\,ms  & 192 & 243\,594 & 45\,110 \\
40\,ms  & 256 & 225\,550 & 63\,154 \\
50\,ms  & 320 & 207\,506 & 81\,198 \\
\bottomrule
\end{tabular}
\end{table}

For preprocessing, each simulation episode is cropped to \(\pm 80\,\text{ms}\) around the fault inception to capture the protection-critical pre-fault steady state and early post-fault transient, while excluding later dynamics that exceed practical decision horizons. Overlapping signal segments are extracted using a sliding window with a 5\,ms stride, matching the typical interrupt cycle of numerical protection relay processors, ensuring the windowing scheme reflects realistic real-time sampling cadences. Window lengths of 10–50\,ms are evaluated, corresponding to 0.5–2.5 fundamental cycles at 50\,Hz and spanning the operationally feasible range for numerical protection relays as defined in IEC~60255-1 and IEEE~C37.114. As summarized in Table~\ref{tab:window_length_overview}, increasing the window length reduces the total number of extracted windows by $\sim$25.8\,\% while raising the share of fault-containing segments from 3.2\,\% to 39.1\,\%, reflecting the trade-off between decision latency and the proportion of informative windows available for training and evaluation.

\begin{table*}[t]
\centering
\caption{Fault classification F1-score across decision horizons.
Mean $\pm$ std over 5 runs; models ordered by average performance.}
\label{tab:fc_decision_horizons}
\begin{tabular}{lccccccc}
\toprule
Window & \bfseries MLP & \bfseries GB & \bfseries KNN & \bfseries RF & \bfseries ET & \bfseries Ridge & \bfseries LG \\
\midrule
50 ms & \textbf{0.990 $\pm$ 0.003} & 0.982 $\pm$ 0.002 & 0.863 $\pm$ 0.003 & 0.838 $\pm$ 0.004 & 0.431 $\pm$ 0.020 & 0.089 $\pm$ 0.002 & 0.071 $\pm$ 0.001 \\
40 ms & 0.978 $\pm$ 0.013 & \textbf{0.982 $\pm$ 0.002} & 0.851 $\pm$ 0.003 & 0.828 $\pm$ 0.004 & 0.300 $\pm$ 0.033 & 0.091 $\pm$ 0.002 & 0.079 $\pm$ 0.001 \\
30 ms & \textbf{0.989 $\pm$ 0.003} & 0.977 $\pm$ 0.002 & 0.831 $\pm$ 0.004 & 0.802 $\pm$ 0.004 & 0.213 $\pm$ 0.008 & 0.093 $\pm$ 0.002 & 0.086 $\pm$ 0.001 \\
20 ms & \textbf{0.993 $\pm$ 0.002} & 0.738 $\pm$ 0.023 & 0.799 $\pm$ 0.006 & 0.771 $\pm$ 0.007 & 0.162 $\pm$ 0.005 & 0.095 $\pm$ 0.003 & 0.090 $\pm$ 0.001 \\
10 ms & \textbf{0.989 $\pm$ 0.003} & 0.418 $\pm$ 0.009 & 0.738 $\pm$ 0.012 & 0.292 $\pm$ 0.026 & 0.098 $\pm$ 0.001 & 0.097 $\pm$ 0.002 & 0.104 $\pm$ 0.004 \\
\bottomrule
\end{tabular}
\end{table*}

\begin{table*}[t]
\centering
\caption{Fault localization MAE (normalized line length) across decision horizons.
Mean $\pm$ std over 5 runs; models ordered by average performance.}
\label{tab:fl_decision_horizons}
\setlength{\tabcolsep}{3pt}
\begin{tabular}{lccccccc}
\toprule
\bfseries Window & \bfseries Stacking & \bfseries MLP & \bfseries Voting & \bfseries GB & \bfseries KNN & \bfseries DT & \bfseries ET \\
\midrule
50 ms & \textbf{9.676 $\pm$ 0.320} & 9.915 $\pm$ 0.331 & 10.623 $\pm$ 0.298 & 14.658 $\pm$ 0.204 & 19.121 $\pm$ 0.138 & 19.714 $\pm$ 0.121 & 21.761 $\pm$ 0.330 \\
40 ms & \textbf{10.086 $\pm$ 0.301} & 10.457 $\pm$ 0.521 & 10.806 $\pm$ 0.152 & 14.590 $\pm$ 0.192 & 19.326 $\pm$ 0.156 & 19.796 $\pm$ 0.113 & 21.723 $\pm$ 0.351 \\
30 ms & \textbf{9.906 $\pm$ 0.286} & 10.179 $\pm$ 0.346 & 10.763 $\pm$ 0.185 & 14.642 $\pm$ 0.182 & 19.646 $\pm$ 0.187 & 20.211 $\pm$ 0.161 & 21.759 $\pm$ 0.359 \\
20 ms & \textbf{9.932 $\pm$ 0.321} & 10.195 $\pm$ 0.372 & 10.820 $\pm$ 0.243 & 14.783 $\pm$ 0.252 & 20.104 $\pm$ 0.165 & 20.977 $\pm$ 0.264 & 21.791 $\pm$ 0.385 \\
10 ms & \textbf{10.304 $\pm$ 0.272} & 10.662 $\pm$ 0.393 & 10.943 $\pm$ 0.140 & 14.646 $\pm$ 0.186 & 20.376 $\pm$ 0.321 & 22.102 $\pm$ 0.481 & 21.730 $\pm$ 0.323 \\
\bottomrule
\end{tabular}
\end{table*}

\subsection{Models, Tasks, and Validation}

Both tasks rely on a common input representation, where three-phase voltage and current signals \(X_{PR}\) from all eight \acp{pr} are concatenated into a multivariate time series, segmented into overlapping sliding windows of varying length (Table~\ref{tab:window_length_overview}).

The \ac{fc} task is formulated as a multi-class classification problem. Each input window receives a label
\begin{equation*}
    y_{\text{FC}} \in \{c_0, c_1, \dots, c_{10}\},
\label{eq:fc_labels}
\end{equation*}
where \(c_0\) denotes ``No Fault” and \(c_1\)--\(c_{10}\) correspond to short-circuit types: SLG (AG, BG, CG), LL (AB, BC, CA), LLG (ABG, BCG, CAG), and the three-phase fault LLL (ABC). Labels are derived from the simulation metadata and attached to all sliding windows. Macro-averaged F1 is reported as the primary metric, accounting for the dominance of the no-fault class in the windowed dataset.

The \ac{fl} task is formulated as a regression task, predicting fault location as a percentage of the line length,

\begin{equation*}
y_{\text{FL}} = \frac{d_{\text{fault}}}{L_{\text{line}}} \cdot 100,
\label{eq:fl_label}
\end{equation*}

where \(d_{\text{fault}}\) is the distance from the sending end and \(L_{\text{line}}\) the total line length. This normalized formulation \([0,100]\%\) supports generalization across different topologies and parallels conventional distance protection, which estimates the fault distance from local V/I signals relative to impedance-based thresholds~\cite{mahr_distanzschutzalgorithmen_2021}. A window is considered only if the fault start is fully contained,

\begin{equation*}
t_{\text{start}} + \epsilon < t_{\textit{fault\_start}} < t_{\text{end}} - \epsilon,
\label{eq:fl_condition}
\end{equation*}

with \(\epsilon = 5\,\mu\text{s}\) ensuring sufficient separation. \ac{fl} performance is assessed by \ac{mae}, which is emphasized as the primary metric due to its direct physical interpretability in terms of normalized line position.

Traditional single-ended impedance locators require scenario-specific compensation for intermediate infeeds and parallel-line currents, which are randomized in this setting; \ac{ml} methods are therefore the appropriate choice for evaluating data-driven performance limits under unknown grid configurations. To probe task-dependent behavior across inductive biases, we evaluate models spanning the families most commonly reported in the protection literature~\cite{oelhaf_scoping_2025}: linear methods (\ac{lg}, \ac{ridge}) as lower-capacity baselines, neighborhood and tree-based models (\ac{knn}, \ac{dt}) representing interpretable nonlinear approaches~\cite{mohanty_decision_2020}, ensemble methods (\ac{et}, \ac{gb}, \ac{rf}, \ac{stacking}, \ac{voting}) reflecting the dominant practical choice in recent studies, and the \ac{mlp} as a higher-capacity feature extractor~\cite{abdullah_ultrafast_2018}. For \ac{fl}, the corresponding regression variants were applied. Models that exceeded a 24\,h training time under the cross-validation protocol were excluded; this affected several classifiers (\ac{bc}, Linear-\ac{svc}, \ac{sgd}, \ac{svc}) and regressors (\ac{svr}, Bagging, \ac{rf}). The reported model sets differ between tasks, reflecting exclusions based on training time limits and task-specific convergence behavior.

All remaining models are implemented in \textit{scikit-learn} with default hyperparameters and standardized inputs (zero mean, unit variance), ensuring a task-agnostic comparison aimed at characterizing task-dependent performance limits rather than optimizing individual models. A 5-fold cross-validation is applied with grouping by simulation episode, preventing information leakage from temporal overlap between adjacent windows and shared fault parameters within an episode. For \ac{fl}, only windows that contain the fault inception time are retained, as localization relies on the initial transient signature; pre-fault and steady-state post-fault windows carry no discriminative spatial content. Centralized access to all relay measurements is assumed throughout, consistent with IEC~61850-based substation communication architectures~\cite{mackiewicz_overview_2006}, establishing a practical performance upper bound.

\section{Experiments and Results}
\label{sec:results}

Table~\ref{tab:fc_decision_horizons} summarizes the \ac{fc} results. Across decision horizons, \ac{fc} performance is largely insensitive to window length for high-capacity models. The \ac{mlp} achieves F1\,$=$\,0.989 at 10\,ms and F1\,$=$\,0.990 at 50\,ms. This indicates that fault-type information is already present in the earliest post-fault transient. \Ac{gb} shows the opposite sensitivity, strong at 30--50\,ms ($\approx$\,0.98) but collapsing to 0.738 and 0.418 at shorter horizons, while \ac{knn} and \ac{rf} degrade gradually with decreasing window length. Linear models fail across all horizons, confirming that \ac{fc} requires nonlinear decision boundaries regardless of available temporal context. The key finding is not that performance improves with longer windows, but that high-capacity nonlinear models solve \ac{fc} from minimal temporal context: the discriminative fault signature is fully present in the 10\,ms transient.

Table~\ref{tab:fl_decision_horizons} shows a starkly different structure for \ac{fl}. The three top-performing models, \ac{stacking}, \ac{mlp}, and \ac{voting}, form a distinct performance tier with MAE values of 9.7--10.9\,\% across all window lengths, varying by less than 0.8\,\% as the window grows from 10\,ms to 50\,ms. A second tier comprising \ac{gb}, \ac{knn}, \ac{dt}, and \ac{et} plateaus at 14--22\,\% and is equally insensitive to window length. Neither performance tier improves meaningfully with additional temporal context, and the gap between them is large and stable across all decision horizons. This two-tier, window-insensitive structure indicates that the localization error floor is not a model capacity problem, as the most expressive models are already in the top tier, but rather is consistent with topology-induced identifiability limits, specifically parallel-line interactions and intermediate infeeds that create ambiguity in the raw \(V/I\) signals regardless of observation duration.

\begin{figure}[ht]
    \centering
    \includegraphics[width=0.95\linewidth]{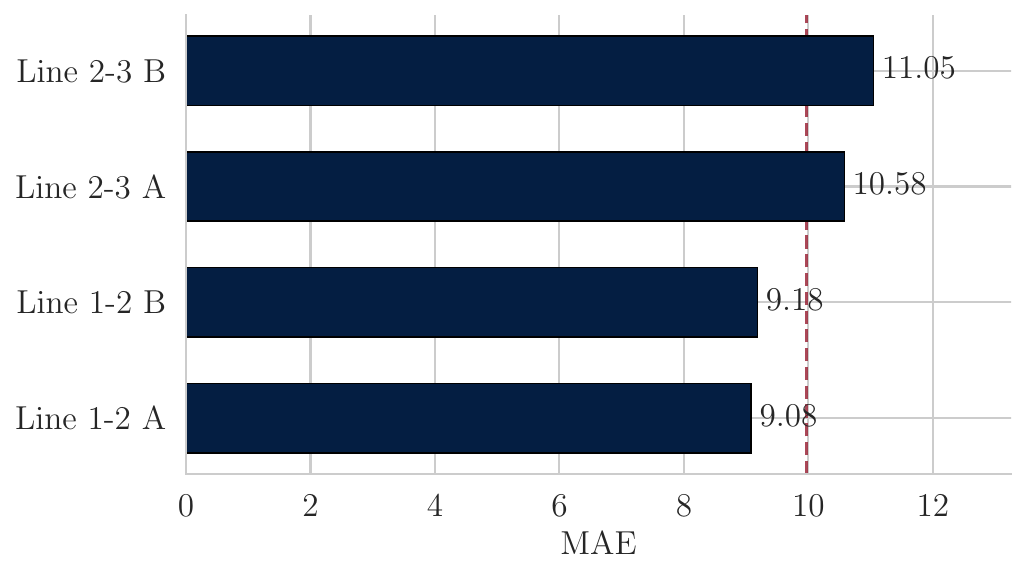}
    \caption{\ac{fl} error by faulted line segment for the MLP model at 50\,ms. The dashed line indicates the mean \ac{mae} across all line segments.}
    \label{fig:fl_mae_by_line}
\end{figure}

Residual \ac{fc} misclassifications are rare across all fault types, consistent with the high aggregate F1 scores in Table~\ref{tab:fc_decision_horizons}. For \ac{fl}, line-resolved error analysis reveals systematic topology-dependent variation: lines subject to stronger parallel-line interactions exhibit higher \ac{mae}, reflecting structural identifiability constraints rather than model deficiencies (see Fig.~\ref{fig:fl_mae_by_line}).

\begin{figure}[ht]
    \centering
    \includegraphics[width=\linewidth]{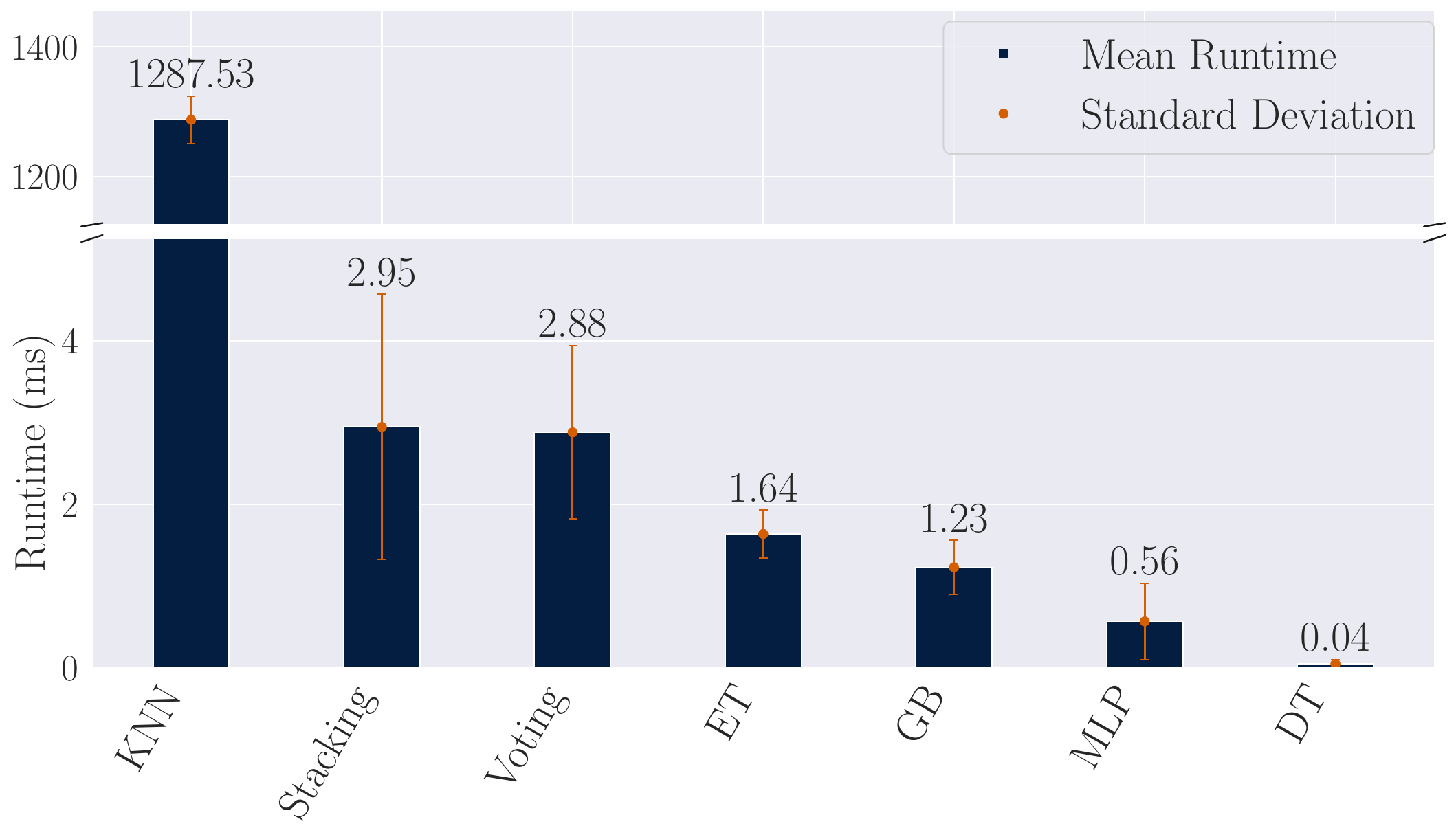}
    \caption{Per-sample inference time of \ac{ml} models on a single CPU core; \ac{knn} shown on a broken axis.}
    \label{overview_runtime_ml_models}
\end{figure}

Inference runtimes for the seven \ac{fl} models are shown in Fig.~\ref{overview_runtime_ml_models}, representing the computationally most demanding setting. Measurements were performed on a 13th Gen Intel\textregistered{} Core\texttrademark{} i7-13700HX CPU using batched single-sample forward passes over 5{,}000 iterations. Reported times reflect inference only. \Ac{dt} achieves the lowest latency ($\approx 0.04\,\mathrm{ms}$), followed by \ac{mlp} ($\approx 0.56\,\mathrm{ms}$) and \ac{gb} ($\approx 1.23\,\mathrm{ms}$). Ensemble aggregation methods \ac{voting} and \ac{stacking} incur higher runtimes ($\approx 2.9\,\mathrm{ms}$), while \ac{knn} exceeds $1\,\mathrm{s}$ and is therefore unsuitable for real-time use. Preprocessing overhead (window extraction and feature standardization) adds less than 0.1\,ms per sample, leaving inference as the dominant cost; the complete pipeline remains within 3.1\,ms for all models except \ac{knn}. All models except \ac{knn} remain well below typical protection decision horizons, indicating that inference time is not the limiting factor for the best-performing nonlinear models in the evaluated centralized setting.

\section{Conclusion}
\label{sec:conclusion}

This paper presented a controlled comparison of machine learning models for \ac{fc} and \ac{fl} under identical sensing, timing, and validation conditions. For \ac{fc}, the high-capacity nonlinear models achieved F1 scores above 0.98 already from 10\,ms decision windows, indicating that fault-type information is available in the earliest transient. For \ac{fl}, the best-performing models reached a stable error of about 10\,\% normalized line length across all evaluated horizons, while weaker models formed a clearly separated second performance tier. Line-resolved analysis showed that localization difficulty depends on the faulted line segment, supporting the interpretation that the remaining error is linked to topology-induced ambiguity rather than insufficient temporal context alone. Overall, the results show that model ranking is strongly task-dependent: high performance on \ac{fc} does not translate automatically into equally strong \ac{fl} performance. These findings provide a controlled reference for future comparisons of learning-based protection methods under shared assumptions. The persistent \ac{fl} error floor suggests that further gains may require additional structural information beyond raw waveform windows, for example through physics-informed approaches that incorporate line parameters or topology knowledge, as recently explored in Graph Neural Network-based fault localization~\cite{kordowich_graph_2025}.

\section*{Acknowledgment}
This project was funded by the Deutsche Forschungsgemeinschaft (DFG, German Research Foundation) - 535389056.

\bibliographystyle{IEEEtran}
\bibliography{refs}

\end{document}